%% file: elsarticle-template-num.tex
 \documentclass[final,3p,times,twocolumn]{elsarticle}

\usepackage{amssymb}
\usepackage{hyperref}
\usepackage{graphicx}
\usepackage{multirow}
\usepackage{xcolor}

\usepackage{amsmath}
\usepackage{amsfonts} 
\usepackage{graphicx}
\usepackage{algorithm}
\usepackage[noend]{algpseudocode} 
\newcommand{\mathds}[1]{\mathbf{1}_{#1}}
\usepackage{booktabs} 
\usepackage[T1]{fontenc}
\usepackage{float}

\journal{Applied Soft Computing}

\begin{document}

\begin{frontmatter}



\title{Ordinal Adaptive Correction: A Data-Centric Approach to Ordinal Image Classification with Noisy Labels}


\author[inst1]{Alireza Sedighi Moghaddam}
\ead{a\_sedighi77@comp.iust.ac.ir}

\author[inst1]{Mohammad Reza Mohammadi\corref{cor1}}
\ead{mrmohammadi@iust.ac.ir}

\affiliation[inst1]{organization={School of Computer Engineering, Iran University of Science and Technology},
            country={Islamic Republic of Iran}}

\cortext[cor1]{Corresponding author}

\begin{abstract}
\input{sections/abstract}
\end{abstract}




\end{frontmatter}




\input{sections/introduction}
\input{sections/related_work}
\input{sections/methodogy}
\input{sections/results}
\input{sections/conclusion}

\appendix


 \bibliographystyle{elsarticle-num} 
 \bibliography{cas-refs}





\end{document}

%% file: sections/abstract.tex
Labeled data is a fundamental component in training supervised deep learning models for computer vision tasks. However, the labeling process, especially for ordinal image classification where class boundaries are often ambiguous, is prone to error and noise. Such label noise can significantly degrade the performance and reliability of machine learning models. This paper addresses the problem of detecting and correcting label noise in ordinal image classification tasks.
To this end, a novel data-centric method called \underline{ORD}inal \underline{A}daptive \underline{C}orrection (ORDAC) is proposed for adaptive correction of noisy labels. The proposed approach leverages the capabilities of Label Distribution Learning (LDL) to model the inherent ambiguity and uncertainty present in ordinal labels. During training, ORDAC dynamically adjusts the mean and standard deviation of the label distribution for each sample. Rather than discarding potentially noisy samples, this approach aims to correct them and make optimal use of the entire training dataset.\\
The effectiveness of the proposed method is evaluated on benchmark datasets for age estimation (Adience) and disease severity detection (Diabetic Retinopathy) under various asymmetric Gaussian noise scenarios. Results show that ORDAC and its extended versions (ORDAC\textsubscript{C} and ORDAC\textsubscript{R}) lead to significant improvements in model performance. For instance, on the Adience dataset with 40\% noise, ORDAC\textsubscript{R} reduced the mean absolute error from \(0.86\) to \(0.62\) and increased the recall metric from \(0.37\) to \(0.49\). The method also demonstrated its effectiveness in correcting intrinsic noise present in the original datasets. This research indicates that adaptive label correction using label distributions is an effective strategy to enhance the robustness and accuracy of ordinal classification models in the presence of noisy data.

%% file: sections/keyword.tex
Label Error Detection \sep Noisy Label Correction \sep Ordinal Classification \sep Label Noise \sep Label Distribution Learning \sep Data-centric Artificial Intelligence

%% file: sections/introduction.tex
\section{Introduction}
\label{sec:intro}

The performance, fairness, and robustness of modern artificial intelligence (AI) systems are fundamentally shaped by the quality of the data they are trained on. This has given rise to a data-centric view of AI, which posits that systematic improvements in data quality are often more impactful than architectural innovations alone \cite{sambasivan2021everyone}. In computer vision, the success of deep neural networks (DNNs) has been fueled by large-scale annotated datasets like ImageNet \cite{russakovsky2015imagenet}. However, creating such datasets is a costly and labor-intensive process. Consequently, researchers often rely on scalable but less controlled methods like crowdsourcing or web scraping, which inevitably introduce noisy or incorrect labels into the training data \cite{cordeiro2020survey}.

The prevalence of noisy labels in benchmark datasets such as ImageNet \cite{russakovsky2015imagenet}, CIFAR-10 \cite{krizhevsky2009learning}, and MNIST \cite{deng2012mnist} presents a significant challenge. DNNs trained on such data can exhibit severely degraded performance, as they may memorize the incorrect annotations. This raises a fundamental question: how can we effectively train models in the presence of label noise? This problem is particularly acute in ordinal classification (or ordinal regression), where the goal is to predict a label from a set of classes with an inherent order, such as estimating age, grading disease severity, or ranking customer satisfaction. In these tasks, the semantic closeness of adjacent classes makes annotation inherently ambiguous and increases the likelihood of label noise.

Existing approaches to mitigate label noise fall into two main categories. Model-centric methods aim to make the learning process itself robust, for example, by designing noise-tolerant loss functions or using regularization to prevent overfitting to incorrect labels \cite{song2022learning}. In contrast, data-centric approaches focus on the data itself, with the dominant strategy being sample selection. Methods like CASSOR \cite{yuan2023cassor} and ICDF \cite{jiang2024noise} identify and filter out samples that are likely mislabeled, training the model on a cleaner subset. While effective, the primary drawback of this approach is the outright removal of samples, which discards the potentially valuable information contained within the features of those instances. This reveals a significant research gap: a need for methods that can correct rather than discard noisy labels by adaptively modeling the uncertainty inherent in ordinal data.

To address this gap, this paper introduces a novel framework named \textbf{ORD}inal \textbf{A}daptive \textbf{C}orrection (\textbf{ORDAC}). Instead of discarding samples suspected of having noisy labels, ORDAC is designed to correct them by leveraging the expressive power of Label Distribution Learning (LDL) \cite{geng2016label}. The core idea is to represent each label not as a single value but as a Gaussian distribution, where the mean represents the label value and the standard deviation quantifies the uncertainty. The ORDAC framework operates iteratively, using the model's own predictions in a cross-validation setup to dynamically update the mean and standard deviation of each training sample's label distribution. This adaptive correction mechanism allows the model to gradually learn from progressively cleaner and more reliable labels, enhancing its robustness and generalization.

The main contributions of this work are:
\begin{itemize}
	\item We propose a novel framework, ORDAC, that shifts the paradigm for handling noisy ordinal labels from sample selection to adaptive label correction.
	\item We introduce a mechanism to dynamically update both the mean and standard deviation of label distributions, allowing the model to explicitly represent and manage uncertainty during training.
	\item We demonstrate through extensive experiments on real-world datasets that our correction-based approach significantly outperforms both standard training methods and state-of-the-art sample selection techniques in the presence of label noise.
\end{itemize}

By focusing on improving data quality at a fundamental level, this research paves the way for more accurate and reliable ordinal classification, particularly in domains where clean data is scarce and label ambiguity is high. Our implementation is publicly available at \url{https://github.com/AlirezaSM/ORDAC/}.

The structure of the paper is as follows: \hyperref[sec:related_work]{Section 2} reviews related works. \hyperref[sec:methodology]{Section 3} describes methodology and the process of noisy label correction is explained. \hyperref[sec:results]{Section 4} is dedicated to evaluating the proposed method, and the experiments conducted to assess its performance are presented. Finally, \hyperref[sec:conclusion]{Section 5} provides the conclusions and potential directions for future work.

%% file: sections/related_work.tex
\section{Related Work}
\label{sec:related_work}

Our work lies at the intersection of noisy-label learning, ordinal classification, and label distribution learning. We survey these areas to highlight the need for methods that go beyond sample filtering toward active label correction for ordinal data.

\subsection{Learning from Noisy Labels}
The paradigm of Data-Centric AI emphasizes that improving data quality is a cornerstone of building robust models \cite{zha2023data}. A central challenge within this paradigm is learning from data with noisy labels. The strategies developed to address this can be broadly grouped into two families.

\textbf{Model-centric} approaches aim to make the learning algorithm itself resilient to noise. This includes designing robust loss functions like Generalized Cross-Entropy \cite{zhang2018generalized} that are less sensitive to large errors, or employing regularization techniques like Mixup \cite{zhang2017mixup} that discourage the model from memorizing incorrect labels. While often effective, these methods treat the dataset as a static, unchangeable entity.

\textbf{Data-centric} approaches, in contrast, focus on improving the dataset itself \cite{cordeiro2020survey}. These methods generally fall into two categories: sample selection and label correction. The more common strategy has been sample selection, which aims to detect and reduce the effect of mislabeled samples. Many early works relied on the small-loss trick, based on the observation that deep networks tend to learn clean samples earlier. This idea underpins methods such as Co-teaching \cite{han2018co}, where two networks exchange small-loss samples, and MentorNet \cite{jiang2018mentornet}, which learns a curriculum to prioritize easier, likely clean samples. Over time, this paradigm has evolved into label quality scoring frameworks. For example, Confident Learning \cite{northcutt2021confident} provides a principled, model-agnostic way to identify label errors by analyzing the relationship between noisy and predicted labels. Similarly, model-agnostic label quality scoring \cite{kuan2022model} introduces multiple scoring metrics to detect low-quality samples, and Dataset Cartography \cite{swayamdipta2020dataset} visualizes learning dynamics to identify hard-to-learn samples that often correspond to mislabeled data. These approaches have proven effective not only in image classification but also in more complex tasks such as object detection \cite{tkachenko2023objectlab} and semantic segmentation \cite{lad2023estimating}.  

In contrast, label correction offers a more direct, data-restorative strategy. Rather than discarding noisy samples, these methods try to infer the correct labels, preserving the full dataset for training. Although promising, this line of work remains less explored compared to sample selection. Some methods adopt a meta-learning framework, where a meta-model learns a correction function for the noisy dataset, as in Meta Label Correction \cite{zheng2021meta}. Others exploit the geometric structure of the feature space, for example, by using graph-based label propagation \cite{zhang2021dualgraph} to refine labels. Despite their strengths, most correction methods are designed for nominal classification tasks. Our work, ORDAC, advances this area by introducing a new label correction mechanism specifically tailored to the unique structural properties of ordinal data.

\subsection{Ordinal Classification and Its Challenges}
Ordinal classification, or ordinal regression, addresses supervised learning tasks where labels possess a natural order, such as age estimation or clinical severity grading \cite{gutierrez2015ordinal}. Ignoring this intrinsic order and treating the problem as nominal classification is suboptimal, as the cost of misclassification is not uniform; for instance, a one-rank error is far less severe than a five-rank error.

To leverage this structure, a diverse set of methodologies has been developed. A prominent family of methods is based on ordinal binary decomposition, which reframes the $K$-rank problem into a series of simpler binary classification tasks (e.g., is the rank $>k$?). Early deep learning approaches like OR-CNN \cite{niu2016ordinal} applied this principle, but a key challenge was ensuring that the binary predictions were monotonically consistent. The CORAL framework \cite{cao2020rank} elegantly solved this by sharing weights across the binary classifiers, a technique further refined by its successor, CORN \cite{shi2023deep}. Other major paradigms include threshold models, which learn a mapping to a latent continuous score that is then partitioned by a set of learned thresholds \cite{vargas2020cumulative, herbrich1999support}, and the direct design of ordinal loss functions that explicitly penalize predictions based on their rank distance from the true label, such as the Weighted Kappa Loss \cite{de2018weighted}. While powerful, these methods are typically designed for clean datasets and their performance can degrade significantly in the presence of label noise, as they lack an explicit mechanism to handle incorrect rank annotations.

\subsection{Label Distribution Learning for Ordinal Tasks}
Label Distribution Learning (LDL) has emerged as a particularly well-suited paradigm for ordinal tasks due to its ability to handle label ambiguity \cite{geng2016label}. Instead of a single hard label, LDL assigns a probability distribution over all possible labels to each instance. For ordinal data, this is often a unimodal distribution (e.g., a Gaussian), which naturally captures the semantic closeness of adjacent classes and the decreasing likelihood of distant ranks.

LDL methods can be categorized as Fixed-form (FLDL) or Adaptive (ALDL) \cite{li2022unimodal}. FLDL methods like DLDL-v2 \cite{gao2018age} assume a static shape for the label distribution (e.g., a Gaussian with a fixed standard deviation). This is a strong assumption that limits the model's capacity to represent varying levels of uncertainty across different samples. ALDL methods, such as those using a Unimodal-Concentrated Loss \cite{li2022unimodal} or modeling ordinal relationships explicitly \cite{wen2023ordinal}, offer more flexibility. However, it is crucial to note that the reported gains of many specialized loss functions have been questioned by studies highlighting the disproportionate impact of evaluation protocols, such as inconsistent data splitting, on final performance \cite{paplham2024call}. While our framework initializes with a fixed-form distribution, characteristic of FLDL, its core novelty lies in a dynamic correction mechanism that adaptively modifies each sample's distribution, thus operating as an ALDL method.

\subsection{Learning Ordinal Regression with Noisy Labels}
The specific challenge of learning ordinal regression from noisy labels is an emerging research frontier. The few existing methods are primarily data-centric and have focused on sample selection. CASSOR \cite{yuan2023cassor}, for example, proposes a class-aware selection strategy that estimates an "insufficiency score" to dynamically adjust the sampling rate for each class. Similarly, ICDF \cite{jiang2024noise} introduces a filtering algorithm based on inter-class feature-space distances to identify and remove noisy samples.

While these state-of-the-art methods demonstrate strong performance, they share a fundamental limitation: they are designed to identify and discard samples. This strategy, by its nature, can lead to the loss of valuable feature information contained in the discarded instances, especially in data-scarce regimes. Furthermore, these methods do not leverage the unique capabilities of LDL to explicitly model label uncertainty as a means to correct, rather than simply remove, noisy labels. Our work is motivated by this clear and critical gap in the literature. We propose a method that uses the principles of LDL not for filtering, but for the adaptive correction of noisy ordinal labels, thereby preserving data while improving its quality throughout the training process.

%% file: sections/methodogy.tex
\section{Methodology}
\label{sec:methodology}

To address the challenge of label noise in ordinal classification, we propose a novel data-centric framework named \textbf{ORD}inal \textbf{A}daptive \textbf{C}orrection (\textbf{ORDAC}). Unlike existing methods that discard potentially noisy samples, ORDAC is designed to correct them by leveraging the principles of Label Distribution Learning (LDL). Our core idea is to represent each ordinal label not as a discrete value, but as a Gaussian distribution characterized by a mean ($\mu$) and a standard deviation ($\sigma$). In this representation, $\mu$ corresponds to the label's value, while $\sigma$ quantifies the model's uncertainty about that label. The framework iteratively refines both $\mu$ and $\sigma$ for each training sample, using the model's own evolving knowledge to dynamically clean the dataset.

\begin{figure*}[htb]
	\centering
	\includegraphics[width=0.9\textwidth]{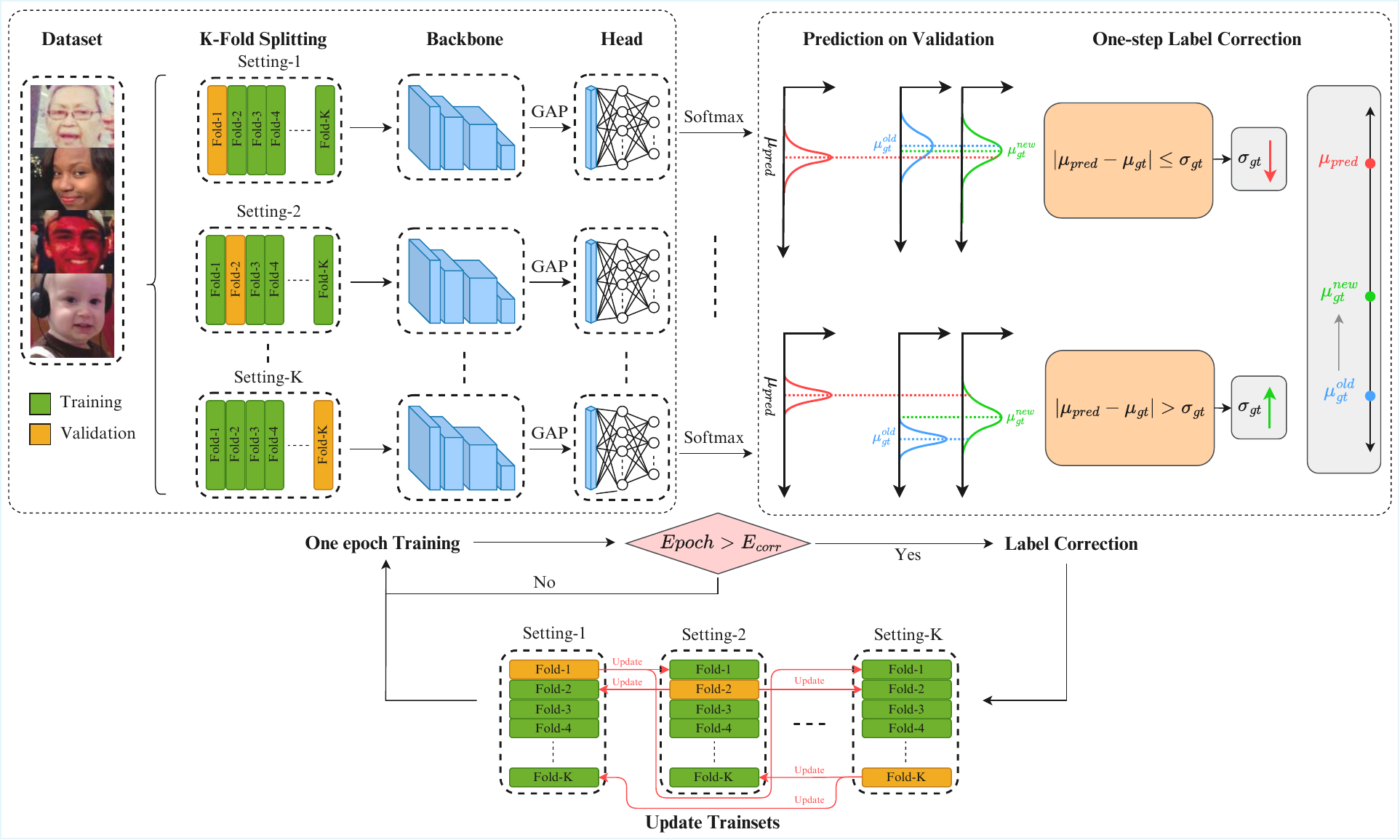}
	\caption{An overview of the proposed ORDAC framework. Using a K-fold setup, models are trained on training folds and make predictions on a validation fold. These predictions are used to correct the label distributions (mean and standard deviation) of the validation samples, which are then propagated back to the training sets for subsequent epochs.}
	\label{fig:method}
\end{figure*}

\subsection{Framework Overview}
The ORDAC framework operates using a cross-training strategy to prevent a model from being biased by its own confident but potentially incorrect predictions on data it has already seen. This ensures that label corrections are always guided by reliable, out-of-sample predictions. The overall workflow, depicted in Figure \ref{fig:method}, consists of the following key stages:

\begin{algorithm*}[htb]
	\caption{The ORDAC Algorithm for Ordinal Adaptive Correction}
	\label{alg:ordac}
	\begin{algorithmic}[1]
		\State \textbf{Input:} Noisy dataset $D = \{(x_i, \tilde{y_i})\}_{i=1}^{N}$, folds $K$, epochs $E_{\text{max}}$, correction start epoch $E_{\text{corr}}$, standard deviation $\sigma_i$
		\State \textbf{Output:} Cleaned dataset $D_{\text{clean}}$, trained models $\{M_k\}_{k=1}^K$
		\State Initialize $K$ data splits $\{D_k\}_{k=1}^K$ and $K$ models $\{M_k\}_{k=1}^K$.
		\State Initialize label distributions $\{\mathcal{N}(\mu_i, \sigma_i^2)\}_{i=1}^N$ with $\mu_i = \tilde{y_i}$.
		\For{$e = 1$ to $E_{\text{max}}$}
		\For{each configuration $k = 1$ to $K$}
		\State Train model $M_k$ on $D_k^{\text{train}}$ for one epoch.
		\EndFor
		\If{$e \geq E_{\text{corr}}$}
		\State Initialize a temporary set for corrected labels $D_{\text{corr}} = \emptyset$.
		\For{each configuration $k = 1$ to $K$}
		\State Get predictions $\hat{Y}_k$ for validation set $D_k^{\text{valid}}$ from model $M_k$.
		\State Compute class-wise means and get shifted predictions $\hat{Y}_{k}^{\text{shifted}}$ using Eq. \ref{eq:pred-shift}.
		\For{each sample $(x_i, \mu_i, \sigma_i) \in D_k^{\text{valid}}$}
		\State Compute correction coefficient $\lambda_i^{\text{corr}}$, update rates $\alpha_i, \beta_i$.
		\State Compute error $e_i = \hat{y}_i^{\text{shifted}} - \mu_i$.
		\State Update $\sigma_i^{\text{new}} = \sigma_i + \alpha_i \times (|e_i| - \sigma_i)$.
		\State Update $\mu_i^{\text{new}} = \mu_i + \beta_i \times e_i$.
		\State Add $(x_i, \mu_i^{\text{new}}, \sigma_i^{\text{new}})$ to $D_{\text{corr}}$.
		\EndFor
		\EndFor
		\State Update all training sets $D_k^{\text{train}}$ with the corrected distributions from $D_{\text{corr}}$.
		\EndIf
		\EndFor
		\State $D_{\text{clean}} = $ union of all final corrected validation folds.
		\State \textbf{Return} $D_{\text{clean}}$, $\{M_k\}_{k=1}^K$.
	\end{algorithmic}
\end{algorithm*}

\begin{enumerate}
	\item \textbf{Data Partitioning and Model Setup:} The training dataset $D$ is split into $K$ folds. We then create $K$ distinct training configurations. For each configuration $k \in \{1, \dots, K\}$, one fold is designated as the validation set ($D_k^{\text{valid}}$), and the remaining $K-1$ folds constitute the training set ($D_k^{\text{train}}$). We initialize $K$ identical Fixed-form LDL (FLDL) models, $M_1, \dots, M_K$, one for each configuration.
	
	\item \textbf{Initialization of Label Distributions:} For a noisy dataset $D = \{(x_i, \tilde{y_i})\}_{i=1}^{N}$, we initialize a Gaussian label distribution $\mathcal{N}(\mu_i, \sigma_i^2)$ for each sample $x_i$. The initial mean $\mu_i$ is set to the provided noisy label $\tilde{y_i}$, and the standard deviation $\sigma_i$ is initialized to a fixed, constant value for all samples, representing uniform initial uncertainty.
	
	\item \textbf{Warm-up Phase:} All $K$ models are trained concurrently on their respective training sets for a specified number of warm-up epochs ($E_{\text{corr}}$). This allows the models to learn initial feature representations and converge to a stable state on the original noisy data before any corrections are made.
	
	\item \textbf{Iterative Correction Phase:} After the warm-up phase, for each subsequent epoch, the adaptive correction mechanism is activated. The label distribution for each sample in the training set is updated based on the predictions from the corresponding model that held it out as a validation sample. For instance, the corrected labels from $D_k^{\text{valid}}$ (generated by model $M_k$) are used to update the corresponding samples in all other training sets $D_{j}^{\text{train}}$ where $j \neq k$. This process is detailed in Section \ref{subsec:correction_mechanism} and Algorithm \ref{alg:ordac}.
\end{enumerate}

\subsection{Adaptive Label Correction Mechanism}
\label{subsec:correction_mechanism}
The core of ORDAC is its two-stage adaptive correction process, which is applied iteratively to each sample in the designated validation fold during the correction phase. This process first corrects for systematic model bias and then performs a sample-specific update of the label distribution.

\subsubsection{Class-wise Prediction Debiasing}
Ordinal regression models are prone to developing a systematic bias towards middle-rank classes. This phenomenon arises from two primary factors. First, the distance-sensitive loss functions commonly used in ordinal regression encourage a conservative model to predict middle-rank classes, as this strategy minimizes the upper bound of the loss in the worst-case scenario. Second, this tendency is often exacerbated by inherent data imbalance, where middle-rank classes frequently contain more samples than those at the extremes of the spectrum. To counteract this and prevent our iterative correction process from collapsing towards a biased mean, we first perform a class-wise debiasing of the model's predictions. For each class $c$, we compute the mean prediction of the model across all samples currently labeled as belonging to that class:
\begin{equation}
	\text{mean}_c = \frac{1}{N_c} \sum_{i=1}^{N} \mathds{1}[\mu_i=c]\hat{y_i}
\end{equation}
where $N_c$ is the number of samples in class $c$, $\mu_i$ is the current mean of the label distribution for sample $i$, $\hat{y_i}$ is the model's predicted mean for sample $i$, and $\mathds{1}[\cdot]$ is the indicator function.

We then shift the prediction for each sample to re-center the class-wise predictions around the true class label. This assumes the label noise is unbiased and does not significantly alter the true mean of each class. The shifted prediction $\hat{y}_{i}^{\text{shifted}}$ for a sample $i$ with label mean $\mu_i$ is:
\begin{equation}
	\hat{y}_{i}^{\text{shifted}} = \hat{y}_i - (\text{mean}_{\mu_i} - \mu_i)
	\label{eq:pred-shift}
\end{equation}

\subsubsection{Sample-wise Distribution Update}
Following the debiasing step, we perform an adaptive update of the mean ($\mu_i$) and standard deviation ($\sigma_i$) for each sample. The magnitude of this update is governed by a sample-specific correction coefficient, $\lambda_i^{\text{corr}}$, which balances model confidence with class frequency:
\begin{equation}
	\lambda_i^{\text{corr}} = \left( \frac{\gamma_i}{1 - \log(\pi_{\mu_i} + \epsilon)} \right)
\end{equation}
Here, $\gamma_i$ is a measure of the model's confidence in its prediction for sample $i$. $\pi_{\mu_i} = N_{\mu_i}/N$ is the prior probability of the sample's current class, which ensures that samples from rare classes are corrected more cautiously. $\epsilon$ is a small constant to prevent numerical instability.

This coefficient scales the base learning rates, $\alpha_{\text{base}}$ and $\beta_{\text{base}}$, to produce sample-specific update rates:
\begin{equation}
	\alpha_i = \alpha_{\text{base}} \times \lambda_i^{\text{corr}}, \quad \beta_i = \beta_{\text{base}} \times \lambda_i^{\text{corr}}
\end{equation}
The prediction error, $e_i$, is calculated using the debiased prediction:
\begin{equation}
	e_i = \hat{y}_i^{\text{shifted}} - \mu_i
\end{equation}
The standard deviation is then updated based on the relationship between the prediction error and the current uncertainty:
\begin{equation}
	\sigma_i^{\text{new}} = \sigma_i + \alpha_i \times (|e_i| - \sigma_i)
\end{equation}
This rule intuitively increases the uncertainty ($\sigma_i$) if the model's error is larger than the current uncertainty, suggesting the label is likely noisy. Conversely, it decreases uncertainty if the error is small, indicating confidence in the label.

Finally, the mean of the label distribution is corrected by shifting it towards the model's prediction:
\begin{equation}
	\mu_i^{\text{new}} = \mu_i + \beta_i \times e_i
\end{equation}
This entire process is summarized in Algorithm \ref{alg:ordac}.

\subsection{Method Variants}
To better understand the behavior of our framework, we introduce two variants designed to isolate specific effects of the correction process:

\begin{itemize}
	\item \textbf{ORDAC\textsubscript{C} (Correct):} This variant is designed to demonstrate that our correction process genuinely improves the overall quality of the dataset. Here, the full iterative ORDAC process is run to generate a static, cleaned dataset. A new model is then trained from scratch on this corrected dataset without any further online corrections. Strong performance from this variant indicates that the corrected labels are of high quality.
	
	\item \textbf{ORDAC\textsubscript{R} (Remove):} This variant explores the impact of removing samples that remain highly uncertain even after correction. It builds on ORDAC\textsubscript{C} by first generating a corrected dataset. It then identifies and removes samples whose uncertainty (standard deviation) failed to decrease from its initial value (i.e., $\sigma_i^{\text{new}} \geq \sigma_i^{\text{initial}}$), treating them as outliers. A new model is then trained on this smaller, filtered-and-corrected dataset. This allows us to study the synergy between label correction and the removal of hard-to-correct samples.
\end{itemize}

%% file: sections/results.tex
\section{Results}
\label{sec:results}

In this section, we present a comprehensive empirical evaluation of our proposed method, ORDAC. We assess its performance on two real-world ordinal classification datasets, one for age estimation and one for medical image analysis. We first detail the experimental setup, including the datasets, evaluation metrics, and noise injection protocol. We then analyze the robustness of ORDAC against controlled label noise, compare it with state-of-the-art methods, and provide in-depth ablation studies to validate our design choices.

\begin{table*}[h!]
	\centering
	\caption{Main results comparing our proposed methods (ORDAC and its variants) against baselines on the Adience and DR datasets under different noise rates ($\tau$). We report Macro-Averaged MAE and Recall (REC). Lower MAE and higher REC are better. Best results are in \textbf{bold}, second best are \underline{underlined}.}
	\label{tab:comparison}
	\resizebox{0.9\textwidth}{!}{%
		\begin{tabular}{@{}lccccccc@{}}
			\toprule
			\textbf{Dataset} & $\tau$ & \textbf{Metric} & \textbf{CORAL \cite{cao2020rank}} & \textbf{DLDL-v2 \cite{gao2018age}} & \textbf{ORDAC} & \textbf{ORDAC\textsubscript{C}} & \textbf{ORDAC\textsubscript{R}} \\ 
			\midrule
			\multirow{6}{*}{Adience} 
			& \multirow{2}{*}{0.0} & MAE & 0.6640 $\pm$ 0.0287 & 0.6343 $\pm$ 0.0505 & 0.5585 $\pm$ 0.0606  & \textbf{0.4929 $\pm$ 0.0296} & \underline{0.5018 $\pm$ 0.0274} \\
			&                      & REC & 0.4669 $\pm$ 0.0135 & 0.5063 $\pm$ 0.0221 & 0.5542 $\pm$ 0.0353  & \textbf{0.5968 $\pm$ 0.0193} & \underline{0.5954 $\pm$ 0.0136} \\ \cmidrule{2-8}
			& \multirow{2}{*}{0.2} & MAE & 0.8349 $\pm$ 0.0289 & 0.7618 $\pm$ 0.0482 & 0.6463 $\pm$ 0.0394  & \underline{0.5642 $\pm$ 0.0400} & \textbf{0.5366 $\pm$ 0.0347} \\
			&                      & REC & 0.3985 $\pm$ 0.0102 & 0.4452 $\pm$ 0.0211 & 0.4988 $\pm$ 0.0226  & \underline{0.5460 $\pm$ 0.0182} & \textbf{0.5624 $\pm$ 0.0185} \\ \cmidrule{2-8}
			& \multirow{2}{*}{0.4} & MAE & 1.0427 $\pm$ 0.0676 & 0.8649 $\pm$ 0.0382 & 0.7192 $\pm$ 0.0393  & \underline{0.6637 $\pm$ 0.0326} & \textbf{0.6283 $\pm$ 0.0379} \\
			&                      & REC & 0.3312 $\pm$ 0.0245 & 0.3775 $\pm$ 0.0290 & 0.4600 $\pm$ 0.0119  & \underline{0.4813 $\pm$ 0.0218} & \textbf{0.4950 $\pm$ 0.0221} \\  
			\midrule
			\multirow{6}{*}{DR} 
			& \multirow{2}{*}{0.0} & MAE & 0.7721 $\pm$ 0.0173 & 0.7324 $\pm$ 0.0268 & \textbf{0.6826 $\pm$ 0.0122}  & 0.7084 $\pm$ 0.0103 & \underline{0.6924 $\pm$ 0.0172} \\
			&                      & REC & 0.4287 $\pm$ 0.0052 & 0.4509 $\pm$ 0.0136 & \textbf{0.4665 $\pm$ 0.0103}  & 0.4559 $\pm$ 0.0058 & \underline{0.4599 $\pm$ 0.0063} \\ \cmidrule{2-8}
			& \multirow{2}{*}{0.2} & MAE & 0.8403 $\pm$ 0.0347 & 0.8025 $\pm$ 0.0260 & \textbf{0.7114 $\pm$ 0.0202}  & 0.7362 $\pm$ 0.0113 & \underline{0.7246 $\pm$ 0.0060} \\
			&                      & REC & 0.3943 $\pm$ 0.0181 & 0.3894 $\pm$ 0.0176 & \textbf{0.4363 $\pm$ 0.0088}  & 0.4208 $\pm$ 0.0033 & \underline{0.4217 $\pm$ 0.0023} \\ \cmidrule{2-8}
			& \multirow{2}{*}{0.4} & MAE & 0.8595 $\pm$ 0.0436 & 0.8488 $\pm$ 0.0493 & 0.8015 $\pm$ 0.0620  & \underline{0.7567 $\pm$ 0.0168} & \textbf{0.7436 $\pm$ 0.0111} \\
			&                      & REC & 0.3702 $\pm$ 0.0189 & 0.3452 $\pm$ 0.0236 & 0.3722 $\pm$ 0.0323  & \underline{0.4168 $\pm$ 0.0057} & \textbf{0.4200 $\pm$ 0.0059} \\
			\bottomrule
		\end{tabular}%
	}
\end{table*}

\subsection{Experimental Setup}

\subsubsection{Datasets}
We use two challenging, publicly available datasets:
\begin{itemize}
	\item \textbf{Adience} \cite{eidinger2014age}: This dataset contains approximately 26K unfiltered face images from Flickr albums, annotated for age and gender. We focus on the age estimation task, which consists of 8 ordinal classes: (0-2), (4-6), (8-13), (15-20), (25-32), (38-43), (48-53), and (60+). The dataset's in-the-wild nature, with variations in lighting, pose, and resolution, makes it a challenging benchmark. Crucially, the labels are derived from user profiles and are known to contain inherent noise due to self-reporting errors or mismatches between the uploader and the person in the photo. We use the aligned version of the dataset and the official 5-fold split as recommended by \cite{paplham2024call}.
	
	\item \textbf{Diabetic Retinopathy (DR)} \cite{diabetic-retinopathy-detection}: This medical imaging dataset contains over 88K high-resolution retinal fundus images for detecting and grading diabetic retinopathy. The task is to classify each image into one of 5 ordinal severity levels: (0: No DR, 1: Mild, 2: Moderate, 3: Severe, 4: Proliferative DR). The dataset exhibits significant challenges, including variations in imaging conditions and image quality. We use the official training/test split and further divide the training data into 5 folds for our cross-validation setup.
\end{itemize}

\subsubsection{Evaluation Metrics}
Given the class imbalance common in ordinal datasets, we use macro-averaged metrics to ensure a fair evaluation across all classes. We report:
\begin{itemize}
	\item \textbf{Macro-Averaged Mean Absolute Error (MAE)}: The primary metric for ordinal regression, measuring the average absolute difference between the predicted and true class ranks.
	\item \textbf{Macro-Averaged Recall (Accuracy)}: The average of per-class recall, measuring the classification accuracy without bias towards majority classes.
\end{itemize}

\begin{figure*}[h!]
	\centering
	\includegraphics[width=0.85\textwidth]{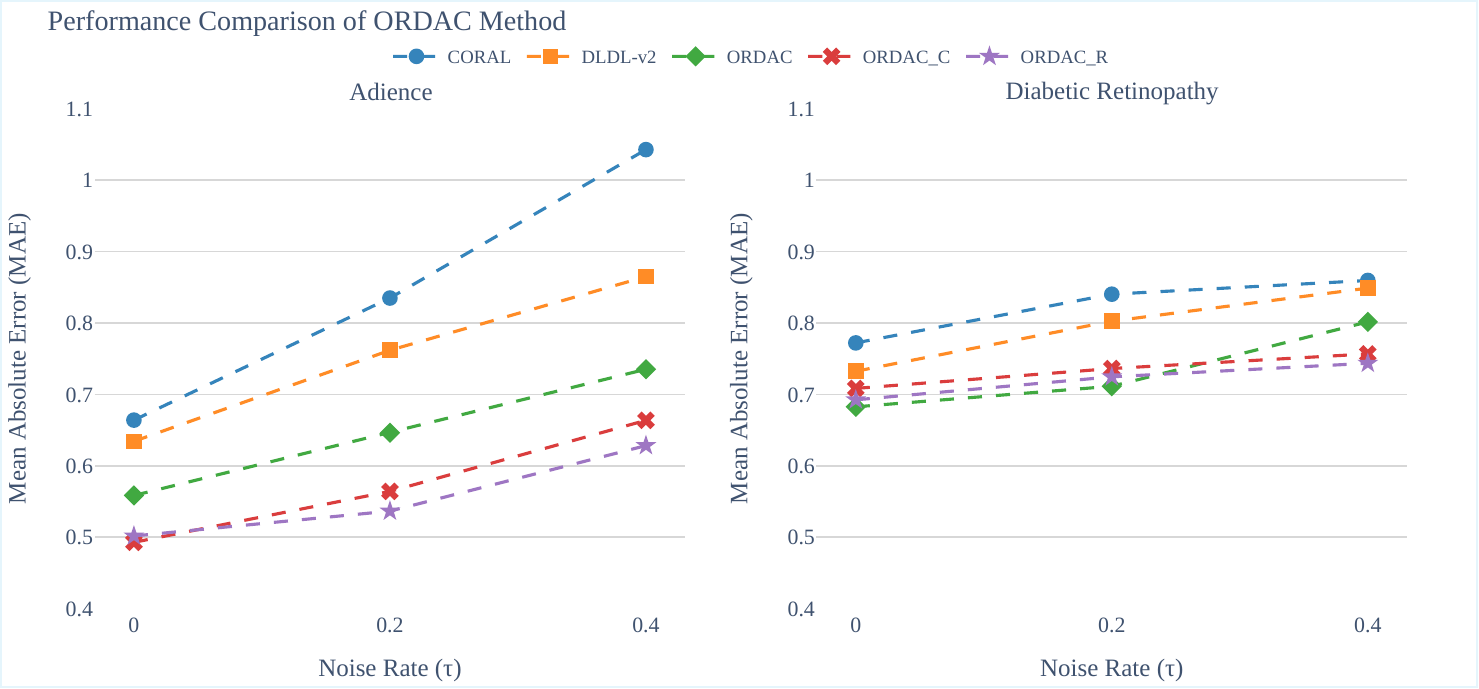}
	\caption{MAE of the proposed methods and baselines on the Adience and DR datasets as a function of the injected noise rate ($\tau$). Lower is better.}
	\label{fig:comparison-mae}
\end{figure*}

\subsubsection{Noise Injection Protocol}
\label{sec:noise-injection-protocol}
Label noise in ordinal tasks is typically asymmetric, as mislabelings are more likely to occur between adjacent, semantically similar classes. To simulate this realistically, we inject \textbf{Gaussian Asymmetric Noise} into the training and validation sets, following the protocol used in prior works \cite{liu2024distributed, yuan2023cassor, jiang2024noise}. A noise transition matrix $T$, where $T_{ij} = P(\tilde{y} = j | y = i)$ is the probability of a true label $i$ being flipped to a noisy label $j$, is generated. The off-diagonal elements are defined by a Gaussian function of the distance between ranks:
\begin{equation}
	T_{ij} \propto \exp\left( -\frac{(i - j)^2}{2\sigma_n^2} \right) \quad \text{for } i \neq j
\end{equation}
where $\sigma_n$ controls the spread of the noise (set to 3 in our experiments). The matrix is then normalized such that the overall noise rate corresponds to a target value $\tau$. The test sets are always kept clean for evaluation.

\begin{figure*}[h!]
	\centering
	\includegraphics[width=\textwidth]{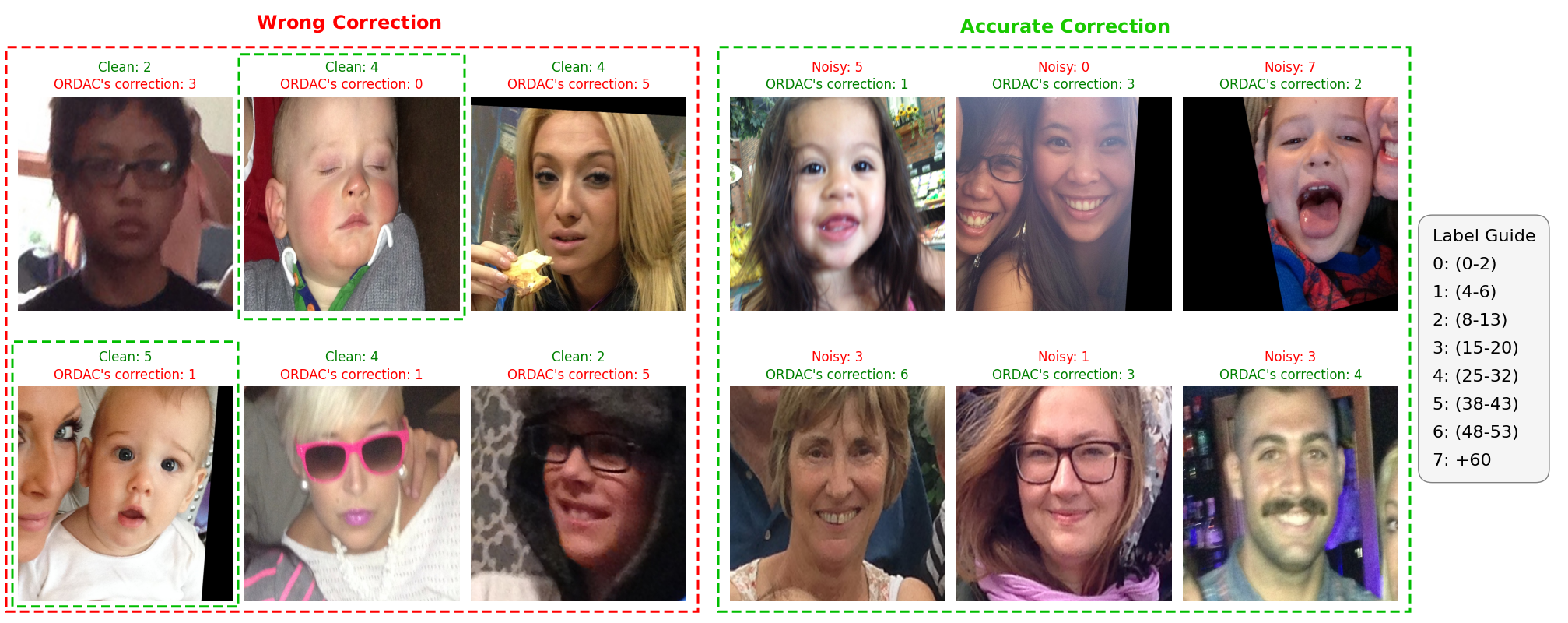}
	\caption{Examples of successful (green box) and unsuccessful (red box) corrections on the Adience dataset with synthetic noise ($\tau=0.4$).}
	\label{fig:top-result-examples}
\end{figure*}

\subsubsection{Implementation Details}
All experiments use a \textbf{ResNet-50} architecture, pre-trained on ImageNet, as the feature extractor backbone. For our proposed method, the model is trained for a maximum of 50 epochs ($E_{\text{max}}$), with the adaptive correction mechanism activating after 10 warm-up periods ($E_{\text{corr}}$). The base learning rates for correction were set to $\alpha_{\text{base}}=0.2$ and $\beta_{\text{base}}=0.8$ based on the hyperparameter analysis in \ref{sec:hyperparameter-analysis}.

\subsection{Performance under Label Noise}
We first evaluate the robustness of ORDAC against synthetic label noise. We compare our three proposed variants (ORDAC, ORDAC\textsubscript{C}, and ORDAC\textsubscript{R}) with two strong baselines: \textbf{CORAL} \cite{cao2020rank}, a standard ordinal regression method, and \textbf{DLDL-v2} \cite{gao2018age}, a fixed-form LDL method. The training and validation sets are corrupted with Gaussian asymmetric noise at rates $\tau \in \{0.2, 0.4\}$. The case $\tau=0$ corresponds to training on the original, uncorrupted datasets.

The results are presented in Table \ref{tab:comparison} and visualized in Figure \ref{fig:comparison-mae}. Our proposed methods consistently and significantly outperform the baselines across both datasets and all noise levels. As expected, the performance of all methods degrades as the noise rate increases. However, the ORDAC framework demonstrates substantially greater resilience. For instance, on Adience with 40\% noise, ORDAC\textsubscript{R} achieves an MAE of 0.6283, a dramatic improvement over CORAL (1.0427) and DLDL-v2 (0.8649).


Interestingly, even with no injected noise ($\tau=0$), our methods still improve performance over the baselines. This strongly suggests that the original datasets contain inherent label noise, which ORDAC successfully identifies and corrects.

\subsection{Analysis of the Correction Mechanism}
To see how our method works in practice, we show several examples of its label corrections in Figure \ref{fig:top-result-examples}. The figure highlights two main outcomes. Successful corrections, where a noisy label was fixed, are marked with green boxes. Apparent errors, where a clean label was changed, are marked with red boxes. 

Crucially, some of these apparent errors (indicated by a small green box) are actually plausible corrections. This happens when the original "clean" label in the dataset was already incorrect. This result shows that our method can fix not only the noise we add for experiments but also the hidden errors that already exist in the dataset.



\begin{figure}[h!]
	\centering
	\includegraphics[width=0.48\textwidth]{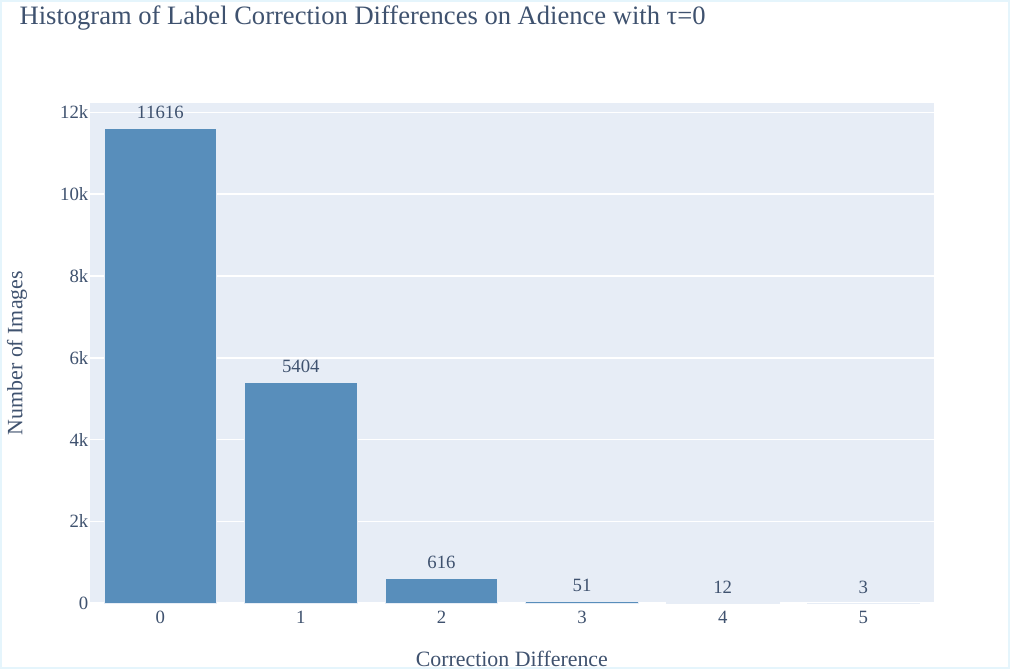}
	\caption{Histogram of label changes made by ORDAC on the original (clean) Adience dataset.}
	\label{fig:correction_hist_adience}
\end{figure}

Quantitatively, we analyze the magnitude of corrections on the original Adience dataset ($\tau=0$). Figure \ref{fig:correction_hist_adience} shows that most corrections are small (a shift of 1 class), with very few large changes, indicating that the model is making targeted, conservative adjustments rather than drastic, random changes.

\begin{table}[h!]
	\centering
	\caption{MAE and RMSE between the true labels and the labels used for training (Noisy for DLDL-v2, Corrected for ORDAC).}
	\label{tab:correction-mae-rmse}
	\resizebox{0.48\textwidth}{!}{%
		\begin{tabular}{@{}ccccc@{}}
			\toprule
			\textbf{Dataset} & \(\tau\) & \textbf{Metric} & \textbf{DLDL-v2 (Noisy)} & \textbf{ORDAC (Corrected)} \\
			\midrule
			\multirow{6}{*}{Adience} & \multirow{2}{*}{0.0} & MAE & 0.0 & 0.4186 \\
			& & RMSE & 0.0 & 0.7116 \\
			\cmidrule{2-5}
			& \multirow{2}{*}{0.2} & MAE & 0.4218 & 0.4933 \\
			& & RMSE & 1.1214 & 0.8070 \\
			\cmidrule{2-5}
			& \multirow{2}{*}{0.4} & MAE & 0.8727 & 0.6481 \\
			& & RMSE & 1.6493 & 0.9894 \\
			\midrule
			\multirow{6}{*}{DR} & \multirow{2}{*}{0.0} & MAE & 0.0 & 0.6106 \\
			& & RMSE & 0.0 & 0.9265 \\
			\cmidrule{2-5}
			& \multirow{2}{*}{0.2} & MAE & 0.3508 & 0.6462 \\
			& & RMSE & 0.8845 & 0.9431 \\
			\cmidrule{2-5}
			& \multirow{2}{*}{0.4} & MAE & 0.6658 & 0.6688 \\
			& & RMSE & 1.2106 & 1.0096 \\
			\bottomrule
		\end{tabular}
	}
\end{table}

\begin{table*}[h!]
	\centering
	\caption{MAE comparison with the CASSOR sample selection method. "Normal Training" uses all noisy data. "ORDAC\textsubscript{C} + CASSOR" applies CASSOR's selection to our corrected dataset.}
	\label{tab:ordac-cassor}
	\resizebox{\textwidth}{!}{
		\begin{tabular}{@{}ccccccc@{}}
			\toprule
			\textbf{Dataset} & \(\tau\) & \textbf{Normal Training} & \textbf{CASSOR} & \textbf{ORDAC\textsubscript{C}} & \textbf{ORDAC\textsubscript{C} + CASSOR} & \textbf{ORDAC\textsubscript{R}}\\
			\midrule
			\multirow{3}{*}{Adience} & 0.0 & 0.5727 $\pm$ 0.0366 & 0.6158 $\pm$ 0.0611 & \underline{0.5038 $\pm$ 0.0458} & 0.5671 $\pm$ 0.0449 & \textbf{0.5033 $\pm$ 0.0353}\\
			& 0.2 & 0.6969 $\pm$ 0.0427 & 0.5902 $\pm$ 0.0442 & \underline{0.5329 $\pm$ 0.0348} & 0.5655 $\pm$ 0.0497 & \textbf{0.5326 $\pm$ 0.0321}\\
			& 0.4 & 0.8200 $\pm$ 0.0632 & \textbf{0.5908 $\pm$ 0.0605} & 0.6321 $\pm$ 0.0288 & \underline{0.5918 $\pm$ 0.0523} & 0.6159 $\pm$ 0.0316 \\
			\midrule
			\multirow{3}{*}{DR} & 0.0 & \underline{0.6712 $\pm$ 0.0040} & \textbf{0.6709 $\pm$ 0.0056} & 0.6893 $\pm$ 0.0107 & 0.7267 $\pm$ 0.0338 & 0.6828 $\pm$ 0.0102 \\
			& 0.2 & 0.7447 $\pm$ 0.0195 & 0.7930 $\pm$ 0.0543 & \underline{0.7283 $\pm$ 0.0100} & 0.7453 $\pm$ 0.0237 & \textbf{0.7149 $\pm$ 0.0086}\\
			& 0.4 & 0.8597 $\pm$ 0.0258 & 0.8240 $\pm$ 0.0736 & \underline{0.7532 $\pm$ 0.0102} & 0.7559 $\pm$ 0.0211 & \textbf{0.7435 $\pm$ 0.0120}\\
			\bottomrule
		\end{tabular}
	}
\end{table*}

We also measure the MAE and RMSE between the true labels and the noisy/corrected labels in Table \ref{tab:correction-mae-rmse}. While ORDAC sometimes increases the MAE of the labels themselves (e.g., for DR at $\tau=0.2$), it consistently improves the final model's performance on the test set (Table \ref{tab:comparison}). This indicates that simply measuring label correctness is insufficient; the ultimate goal is to produce a more generalizable model, which ORDAC achieves. The improved model performance suggests that even when a label is not corrected perfectly, it is often moved closer to the true value, or its associated uncertainty is adjusted appropriately, leading to better overall training dynamics.

%

\subsection{Comparison with State-of-the-Art Sample Selection}
We compare our correction-based approach with \textbf{CASSOR} \cite{yuan2023cassor}, a state-of-the-art sample selection method for noisy ordinal regression. To ensure a fair comparison, we use their official implementation and train it with the same ResNet-50 backbone. We evaluate training with CASSOR's selected clean subset versus training with our corrected labels (ORDAC\textsubscript{C} and ORDAC\textsubscript{R}).

The results in Table \ref{tab:ordac-cassor} show that our correction-based methods generally outperform sample selection, especially as the noise rate increases. On the DR dataset with 40\% noise, ORDAC\textsubscript{R} achieves an MAE of 0.7435, significantly better than CASSOR's 0.8240. This suggests that correcting noisy labels is a more data-efficient strategy than simply discarding them. An interesting case is Adience at $\tau=0.4$, where CASSOR performs best. This may indicate that at very high noise levels, aggressive filtering can be more effective than attempting to correct highly unreliable labels. We also test a hybrid approach (ORDAC\textsubscript{C} + CASSOR), which yields strong results, suggesting that correction and selection are complementary.

\subsection{Ablation Studies}

\subsubsection{Effect of Class-wise Prediction Debiasing}
\begin{figure*}[h!]
	\centering
	\includegraphics[width=\textwidth]{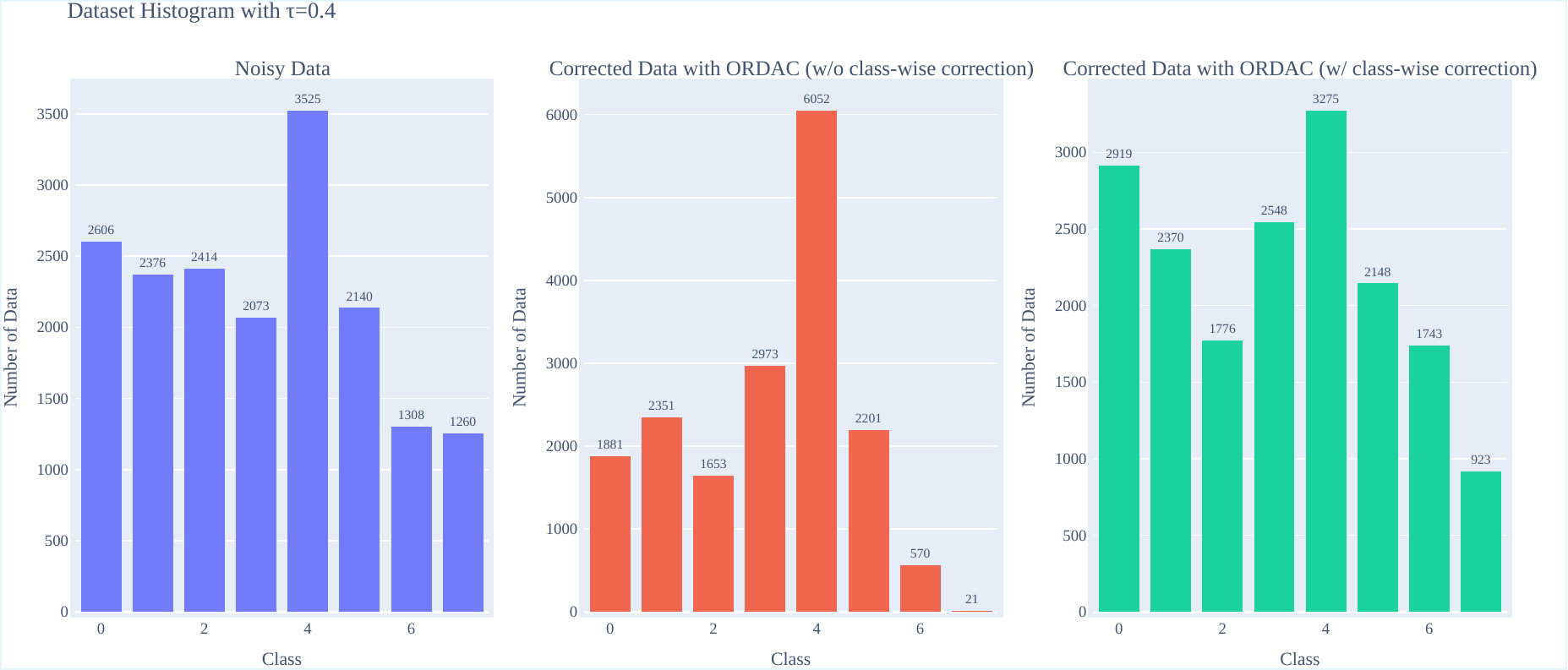}
	\caption{Number of samples per class before and after correction on Adience ($\tau=0.4$), with and without the class-wise debiasing step. Debiasing prevents a collapse into the majority class.}
	\label{fig:class-wise-data-histogram}
\end{figure*}

We evaluate the importance of the class-wise debiasing step described in Section \ref{subsec:correction_mechanism}. Table \ref{tab:class-wise-correction-effect} shows that removing this step significantly degrades performance, especially at higher noise rates. As shown in Figure \ref{fig:class-wise-data-histogram}, without debiasing, the correction process develops a strong bias towards the majority middle class, drastically depleting the minority classes at the ends of the spectrum. The debiasing step successfully mitigates this, ensuring a more balanced and accurate correction process.

\begin{table}[h!]
	\centering
	\caption{Effect of removing the class-wise prediction debiasing step on the Adience dataset.}
	\label{tab:class-wise-correction-effect}
	\resizebox{0.48\textwidth}{!}{%
		\begin{tabular}{@{}cccc@{}}
			\toprule
			\(\tau\) & \textbf{DLDL-v2} & \textbf{ORDAC (w/o debiasing)} & \textbf{ORDAC (w/ debiasing)}\\
			\midrule
			0.0 & 0.6343 $\pm$ 0.0505 & 0.6033 $\pm$ 0.0362 & \textbf{0.5585 $\pm$ 0.0606} \\
			0.2 & 0.7618 $\pm$ 0.0482 & 0.6960 $\pm$ 0.0893 & \textbf{0.6463 $\pm$ 0.0394} \\
			0.4 & 0.8649 $\pm$ 0.0382 & 0.8366 $\pm$ 0.0516 & \textbf{0.7192 $\pm$ 0.0393} \\
			\bottomrule
		\end{tabular}
	}
\end{table}

\subsubsection{Hyperparameter Analysis}
\label{sec:hyperparameter-analysis}
We analyzed the sensitivity to the base correction rates, $\alpha_{\text{base}}$ and $\beta_{\text{base}}$, on Adience with $\tau=0.4$. The results in Table \ref{tab:hyper-tuning} show that the best performance is achieved with a small $\alpha_{\text{base}}$ (0.2) and a large $\beta_{\text{base}}$ (0.8). This aligns with our intuition: the model should be aggressive in correcting the label's mean value ($\mu$) but conservative and stable when updating its uncertainty ($\sigma$). We also found that an initial standard deviation ($std_{\text{init}}$) of 0.75 yielded the best results, providing a good balance for initial uncertainty representation.

\begin{table}[h!]
	\centering
	\caption{Hyperparameter tuning results (MAE) for $\alpha_{\text{base}}$ and $\beta_{\text{base}}$ on Adience ($\tau=0.4$).}
	\label{tab:hyper-tuning}
	\resizebox{0.3\textwidth}{!}{%
		\begin{tabular}{@{}cccc@{}}
			\toprule
			\(\tau\) & \(\alpha_{\text{base}}\) & \(\beta_{\text{base}}\) & \textbf{ORDAC} \\
			\midrule
			\multirow{9}{*}{0.4} & \multirow{3}{*}{0.2} & 0.2 & 0.8773 $\pm$ 0.0399\\
			& & 0.5 & 0.8269 $\pm$ 0.0572 \\
			& & 0.8 & \textbf{0.7192 $\pm$ 0.0393}\\
			\cmidrule{2-4}
			& \multirow{3}{*}{0.5} & 0.2 & 0.8807 $\pm$ 0.0331\\
			& & 0.5 & 0.7983 $\pm$ 0.0515\\
			& & 0.8 & 0.7330 $\pm$ 0.0581\\
			\cmidrule{2-4}
			& \multirow{3}{*}{0.8} & 0.2 & 0.8760 $\pm$ 0.0350\\
			& & 0.5 & 0.8147 $\pm$ 0.0681\\
			& & 0.8 & 0.7463 $\pm$ 0.0897\\
			\bottomrule
		\end{tabular}
	}
\end{table}

%% file: sections/conclusion.tex
\section{Conclusion}
\label{sec:conclusion}

In this paper, we addressed the critical challenge of label noise in ordinal classification, a problem that undermines model performance, particularly in domains with inherent annotation ambiguity. We argued that the dominant data-centric paradigm of sample selection, which discards potentially noisy instances, is suboptimal as it leads to the loss of valuable data. To overcome this limitation, we introduced ORDAC, a novel framework that shifts the focus from sample removal to adaptive label correction. By representing each label as a dynamic Gaussian distribution, our method successfully leverages the model's own evolving knowledge to iteratively refine both the value (mean) and uncertainty (standard deviation) of labels in the training set.

Our extensive experiments on two real-world datasets, Adience and Diabetic Retinopathy, demonstrated the effectiveness of this corrective approach. The results conclusively show that ORDAC and its variants significantly outperform standard ordinal regression baselines and state-of-the-art sample selection methods, especially under high levels of asymmetric label noise. Crucially, we also found that ORDAC improves performance even on the original, uncorrupted datasets, providing strong evidence of its ability to identify and correct inherent, real-world label errors. Our analysis confirmed that this performance gain is driven by a robust mechanism that not only moves noisy labels closer to their true values but also intelligently manages label uncertainty throughout the training process.

For future work, several promising avenues exist. While our method uses a Gaussian distribution, exploring more flexible, non-parametric distributions could allow for modeling more complex types of label ambiguity. Furthermore, developing a mechanism to automatically tune the correction rates based on an online estimation of the dataset's noise level would enhance the framework's autonomy. Finally, applying the principles of ORDAC to other tasks and domains presents an exciting direction for future research.